\documentclass{article}

\usepackage{microtype}
\usepackage{graphicx}
\usepackage{subfigure}
\usepackage{booktabs} % for professional tables
\usepackage{amsmath}

\usepackage{hyperref}

\usepackage[accepted]{icml2020}

\icmltitlerunning{ICML 2020 Workshop on Machine Learning for Global Health}

\begin{document}

\twocolumn[
\icmltitle{Prediction of neonatal mortality in Sub-Saharan African countries using data-level linkage of multiple surveys}



\icmlsetsymbol{equal}{*}

\begin{icmlauthorlist}
\icmlauthor{Girmaw Abebe Tadesse}{1}
 \icmlauthor{Celia Cintas}{1}
 \icmlauthor{Skyler Speakman}{1}
 \icmlauthor{Komminist Weldemariam}{1}
% \icmlauthor{Iona Millwood}{3}
% \icmlauthor{Zhengming Chen}{3}
% \icmlauthor{Tingting Zhu}{2}
% \icmlauthor{Buiui Eueu}{ed}
% \icmlauthor{Aeuia Zzzz}{ed}
% \icmlauthor{Bieea C.~Yyyy}{to,goo}
% \icmlauthor{Teoau Xxxx}{ed}
% \icmlauthor{Eee Pppp}{ed}
\end{icmlauthorlist}

%  \author{$^{1,2}$, $^{2}$, $^{2}$, $^{2}$, $^{2}$,\\ \textbf{$^{2}$, $^{2}$\vspace{0.5cm} }\\
% $^{1}$IBM Research - Africa, Nairobi, Kenya \hspace{2cm}
% $^{2}$ University of Oxford, UK
% }

\icmlaffiliation{1}{IBM Researh - Africa, Nairobi, Kenya}
% \icmlaffiliation{2}{Institute of Biomedical Engineering, University of Oxford, UK}
% \icmlaffiliation{3}{Nuffield Department of Population Health, University of Oxford, UK}

\icmlcorrespondingauthor{Girmaw Abebe Tadesse}{\textcolor{blue}{girmaw.abebe.tadesse@ibm.com}}
% \icmlcorrespondingauthor{Tingting Zhu}{tingting.zhu@eng.ox.ac.uk}

\icmlkeywords{Machine Learning, ICML}

\vskip 0.3in
]



\printAffiliationsAndNotice{}  % leave blank if no need to mention equal contribution

\begin{abstract}
Existing datasets available to address crucial problems, such as child mortality and family planning discontinuation in developing countries, are not ample for data-driven approaches. This is partly due to disjoint data collection efforts employed across locations, times, and variations of modalities. On the other hand, state-of-the-art methods for small data problem are confined to image modalities. In this work, we proposed a data-level linkage of disjoint surveys across Sub-Saharan African countries to improve prediction performance of neonatal death and provide cross-domain explainability.
\end{abstract}

\section{Introduction}~\label{sec:introduction}
%\vspace{-0.4cm}
 Different surveying efforts were conducted to understand the global health challenges in developing countries, which include Demographic and Health Surveys (DHS) Program~ \cite{dhs}, Knowledge Integration (KI) Data~\cite{ki}, and Performance Monitoring for Action (PMA2020)~\cite{pma}. However, these surveys are often utilised in silos with minimal intra- and inter-country integration. Thus, effective utilisation of small but multi-domain data is  beneficial to address problem domains known for data scarcity, or when more data collection is not economically feasible.  

State-of-the-art methods for small data challenges use data-augmentation~\cite{salamon2017deep}, generation~\cite{douzas2018effective} and transfer learning~\cite{sung2018learning} techniques. Though augmentation and generation help to create artificial samples, the intelligibility of these samples is still limited by the small data size in order to generate samples with enough variance.   Transfer learning is the most extensively explored solution for small data problems in the existing literature, and its common strategies include  multitask learning~\cite{tschandl2018ham10000}, few-shot learning ~\cite{guo2019new}, domain adaptation~\cite{pan2009survey}. These existing methods are often applied during or after modelling, and data-level linkage is not well exploited. The critical limitation of existing solutions 
%for small and/or multi-source data problems
is their confinement to image modality and the assumed availability of data-rich source domain, which may not be the case in addressing pressing global health challenges, e.g. neonatal death using survey data. 
%A few existing works exist on cross-domain relation discovery~\cite{kim2017learning} and record linkage~\cite{boratto2019exploring}.
%, e.g., to remove replicated samples or creating longitudinal profiles of individuals~\cite{sayers2016probabilistic}. However, these linkages assume the availability of common attributes across datasets or domains. 
%
%  In addition,  transfer learning approaches for small data cases often assume the availability of data-rich source domains, and the transfer of knowledge is one-directional, i.e., source to target. The linkage of multiple-domains through transfer-learning is applied far from the data level and hence  its interpretation is not straightforward. Moreover, existing data-level linkage approaches require common attributes across the datasets to make the link-up.

In this work, we propose a principled approach to link disjoint datasets, which do not have overlapping of either their samples (rows) or features (columns) but related to the same outcome of interest. Examples of such datasets include PMA collected in Ethiopia in 2016 and DHS collected in Ghana in 2014, which both can be used to understand neonatal death.

\section{Methods} 
\begin{figure*}
    \centering
    \includegraphics[width=0.75\linewidth]{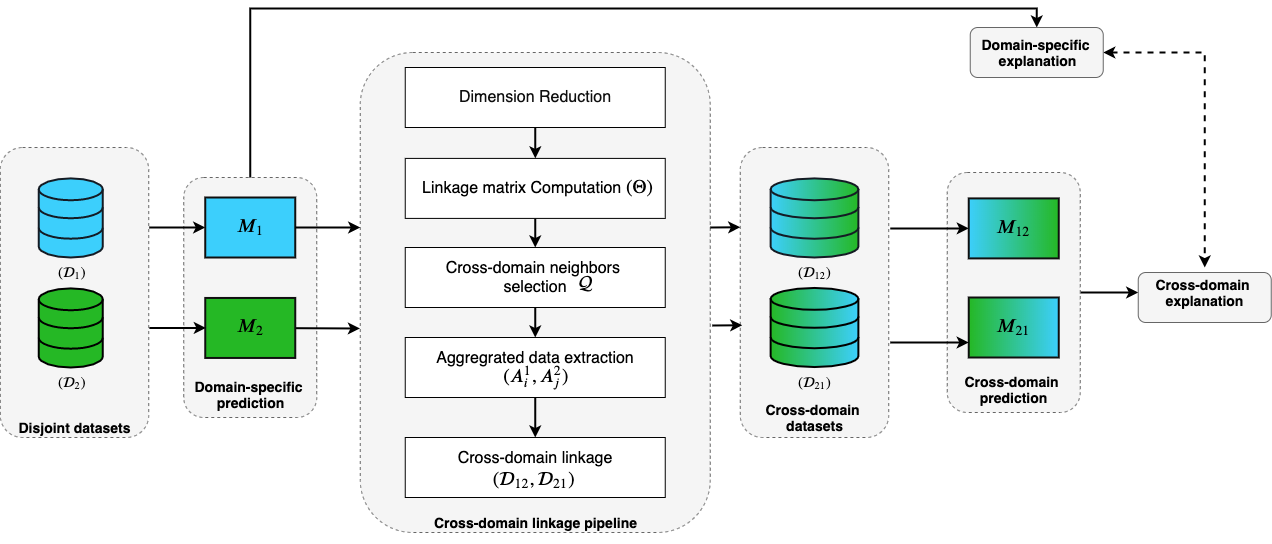}
    \caption{Block diagram of the proposed data-level cross-domain linkage of disjoint datasets.}% Dimension reduction helps to project the datasets into equi-dimensional feature spaces, upon which  datasets are linked.}
    \label{fig:bdg}
\end{figure*}

%\vspace{-0.1cm}
% \paragraph{Problem formulation}
% Given two disjoint datasets, $\mathcal{D}_1=(S^1_i)_{i=1}^N$ and $\mathcal{D}_2=(S^2_j)_{j=1}^M$, where $S^1_i=(f^1_{i1}, f^1_{i2}, \cdots, f^1_{iK})$ and $S^2_j=(f^2_{j1}, f^2_{j2}, \cdots, f^2_{jL)}$, and $N$ and $M$ represent the numbers of samples in $\mathcal{D}_1$ and $\mathcal{D}_2$, respectively;  design a data-level linkage to augment their feature spaces from $\mathbb{R}^K$ and $\mathbb{R}^L$ in $\mathcal{D}_1$ and $\mathcal{D}_2$ into $\mathbb{R}^{K+L}$ in linked $\mathcal{D}_{12}$ or $\mathcal{D}_{21}$,  in order to improve the prediction performance on each of $\mathcal{D}_1$ and $\mathcal{D}_2$.
%
% \paragraph{Disjoint datasets}
% The two disjoint datasets are $\mathcal{D}_1=(S^1_i)_{i=1}^N$ and $\mathcal{D}_2=(S^2_j)_{j=1}^M$, where $N$ and $M$ represent the numbers of samples in $\mathcal{D}_1$ and $\mathcal{D}_2$, respectively. $S^1_i$ and $S^2_j$ in  $\mathcal{D}_1$ and $\mathcal{D}_2$ are  represented by different feature types and dimensions, $K$ and $L$, respectively, i.e., $.
%
% \paragraph{Domain-specific prediction}
% First, domain-specific modelling is applied per dataset, $f^1_\theta~:~\mathcal{D}_1~\rightarrow~y^1$ and $f^2_\theta~:~\mathcal{D}_2~\rightarrow~y^2$, where $y^1$ and $y^2$ are the ground truth data for $\mathcal{D}_1$ and $\mathcal{D}_2$, respectively.  In addition, important features could be extracted from $f^1_\theta$ and $f^2_\theta$ for the feature-importance-based linkage later.
%
%\paragraph{Cross-domain linkage pipeline}
%\vspace{-0.2cm}
 The proposed framework (see Fig.~\ref{fig:bdg}) involves the linkage of samples across two disjoint datasets: $\mathcal{D}_1=(S^1_i)_{i=1}^N$ and $\mathcal{D}_2=(S^2_j)_{j=1}^M$, where $S^1_i=(f^1_{i1}, f^1_{i2}, \cdots, f^1_{iK})$ and $S^2_j=(f^2_{j1}, f^2_{j2}, \cdots, f^2_{jL)}$, and $N$ and $M$ represent the numbers of samples; $K$ and $L$ represent the number of features in $\mathcal{D}_1$ and $\mathcal{D}_2$, respectively. Since the datasets are disjoint and have different feature spaces, we, first, apply dimension reduction to project $\mathcal{D}_1$ and $\mathcal{D}_2$ into equal-dimensional feature spaces. To do so, three techniques are employed that use: \textit{feature-importance}, \textit{principal component analysis (PCA)}, and \textit{autoencoder (AE)}.
 
Feature-importance-based linkage utilises the t-score of a feature in each dataset after domain-specific modelling. %Given $f^1_\theta$ and $f^2_\theta$, we group and sort the features in each dataset based on t-score direction and values.  
%Example for $\mathcal{D}_1$, the t-score values of its features  are arranged as  $\mathbf{t}^1=\{\mathbf{t}^1_+, \mathbf{t}^1_-\}$, where $\mathbf{t}^1_+ =(f^1_i)_{i=1}^{p_1}$ and $\mathbf{t}^1_-=(f^1_i)_{i=1}^{n_1}$ represent the the number of positive and negative directed features, respectively,  $p_1+n_1=K$. Similarly, the features are grouped and sorted for $\mathcal{D}_2$, resulting $\mathbf{t}^2=\{\mathbf{t}^2_+, \mathbf{t}^2_-\}$, $\mathbf{t}^2_+ =(f^2_i)_{i=1}^{p_2}$, $\mathbf{t}^2_- =(f^2_i)_{i=1}^{n_2}$, $p_2+n_2=L$.
We obtain the minimum number of positive ($p_{min}$) and negative ($n_{min}$) features across the two datasets, and represent the samples $\mathcal{D}_1$ and  $\mathcal{D}_2$, using the top $p_{min}$ and $n_{min}$ features sorted in descending order.
%By doing this, we reduce the feature dimension from $K$ in $\mathcal{D}_1$ and from $L$ in $\mathcal{D}_2$ to $p_{min}+n_{min} = R$ in $\mathcal{\hat{D}}_1$ and $\mathcal{\hat{D}}_2$. Thus,  $\mathcal{\hat{D}}_1= (\hat{S^1_i})_{i=1}^N$ and $\hat{S^1_i}=(\hat{f}^1_{i1}, \hat{f}^1_{i2}, \cdots, \hat{f}^1_{iR})$, where $\hat{f}^1_{i1}$ and $\hat{f}^1_{iR}$ represent the top positive and negative features in $\mathcal{D}_1$, respectively. Similarly, $\mathcal{\hat{D}}_2= (\hat{S^2_j})_{j=1}^M$ and $\hat{S^2_j}=(\hat{f}^2_{j1}, \hat{f}^2_{j2}, \cdots, \hat{f}^2_{jR})$. 
PCA-based linkage applies principal component analysis  of $\mathcal{D}_1$ and $\mathcal{D}_2$ followed by projections using the top $R$ Eigen vectors. 
%resulting are then selected to project $\mathcal{D}_1$ and $\mathcal{D}_2$ into $R$-dimensional feature space, $\mathcal{\hat{D}}_1=\mathcal{D}_1^{NxK}*P_1^{KxR}$ and $\mathcal{\hat{D}}_2=\mathcal{D}_2^{MxK}*P_2^{MxR}$.
%,  which also reduce the feature dimensions from $K$ and $L$ to $R$, respectively.  
Similarly, autoencoder-based linkage employs dense encoders and decoders trained on each dataset separately, and the output of the encoder part ($\mathcal{E}(\cdot)$) is treated as a  latent space with reduced dimension. The layer size of the encoder outputs in these two autoencoders is set to be $R$-dimensional. Any of the dimension reduction techniques results in $R$-dimensional  $\mathcal{\hat{D}}_1$ and $\mathcal{\hat{D}}_2$ from $\mathcal{D}_1$ and $\mathcal{D}_2$, respectively.  The  linkage matrix is obtained by applying Euclidean based distance computation for every pair of samples in $\mathcal{D}_1$ and $\mathcal{D}_2$. Close neighbours for a sample are then identified in the other dataset from the distance metric, and the feature values of the neighbors are the median-aggregated and concatenated with the original feature space of a sample, resulting linked datasets $\mathcal{D}_{12}$ and $\mathcal{D}_{21}$. 
 \section{Experiments}
\begin{figure}[t]
    \centering
    \includegraphics[width=0.3\linewidth]{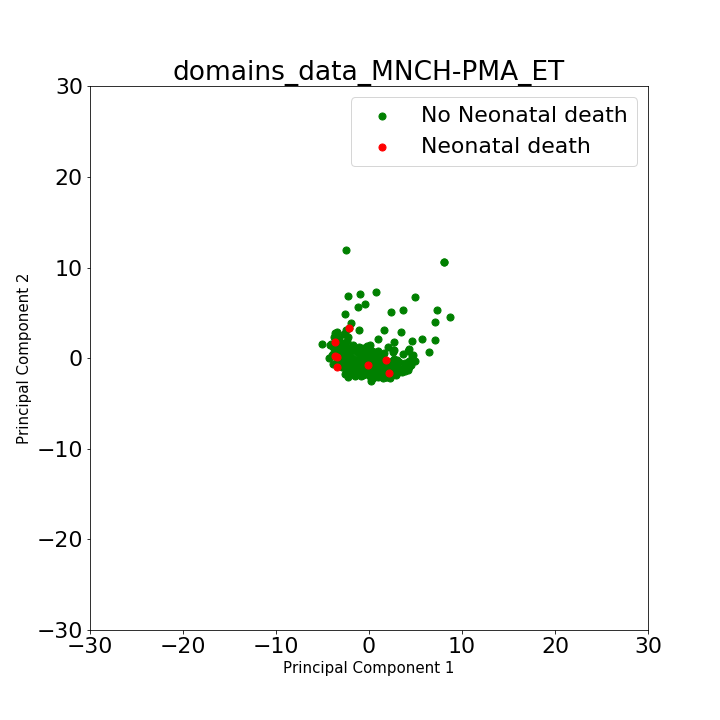}
    \includegraphics[width=0.3\linewidth]{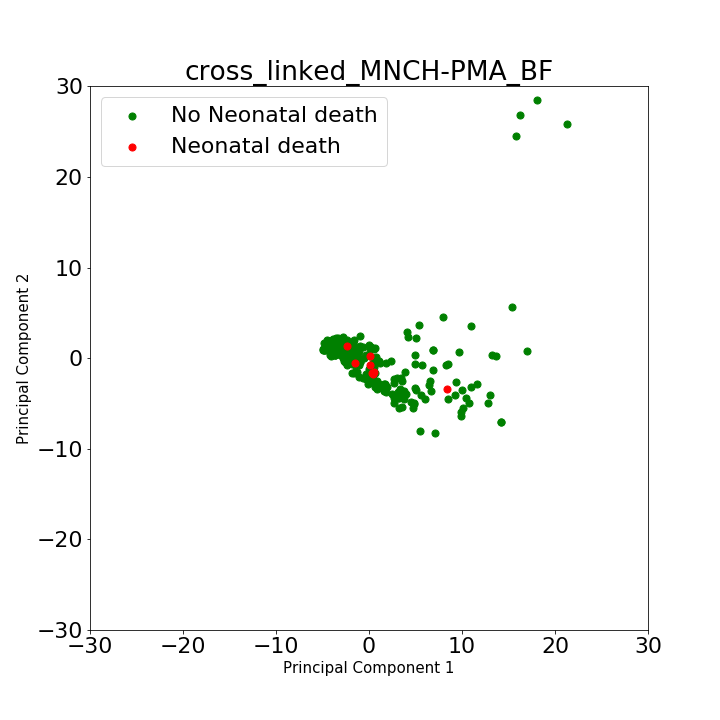}
    \includegraphics[width=0.3\linewidth]{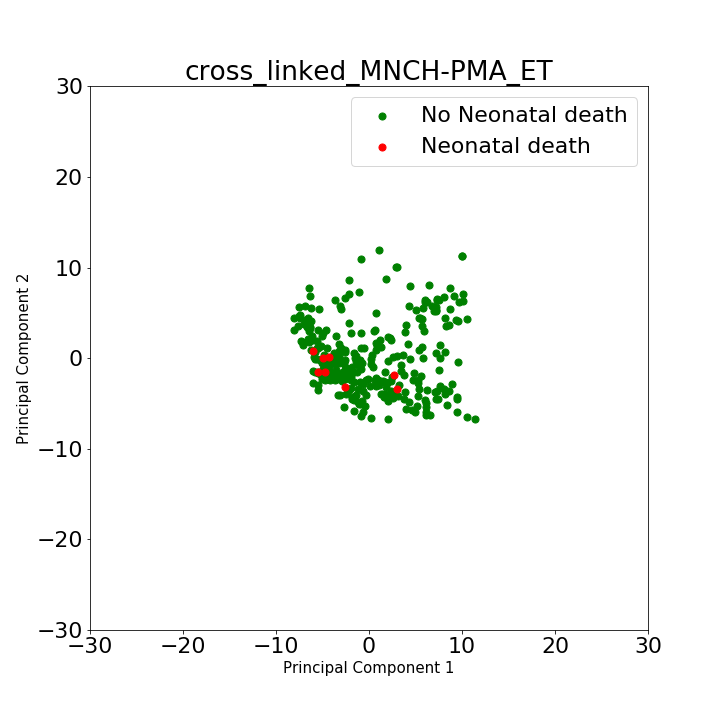}
    \includegraphics[width=0.3\linewidth]{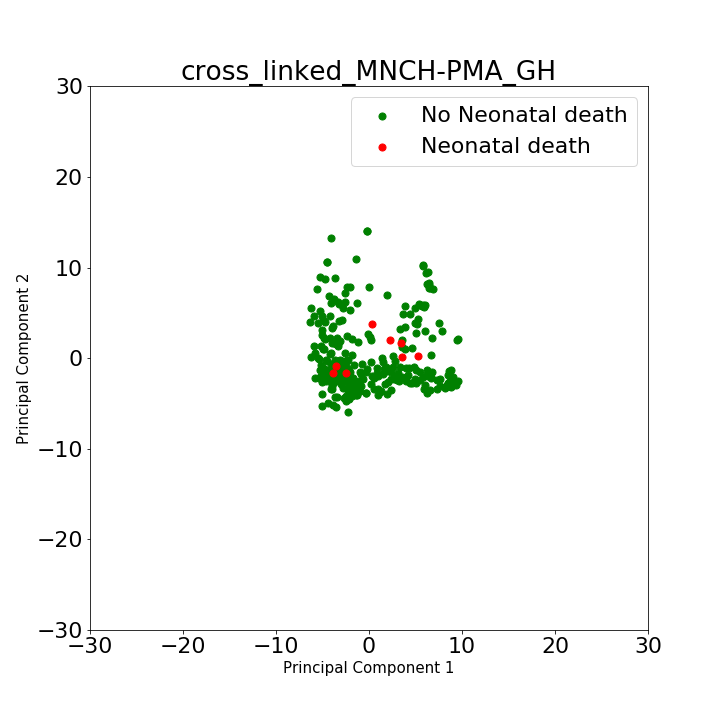}
    \includegraphics[width=0.3\linewidth]{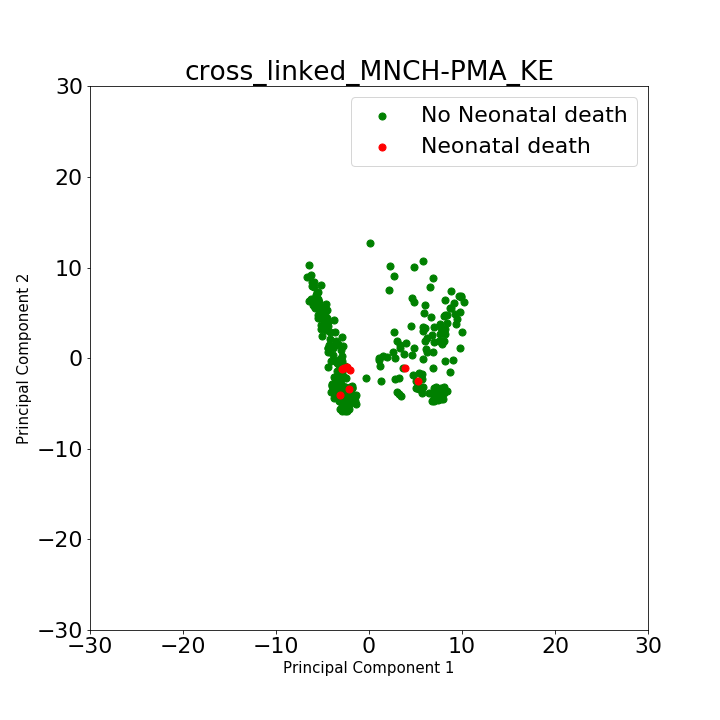}
    \includegraphics[width=0.3\linewidth]{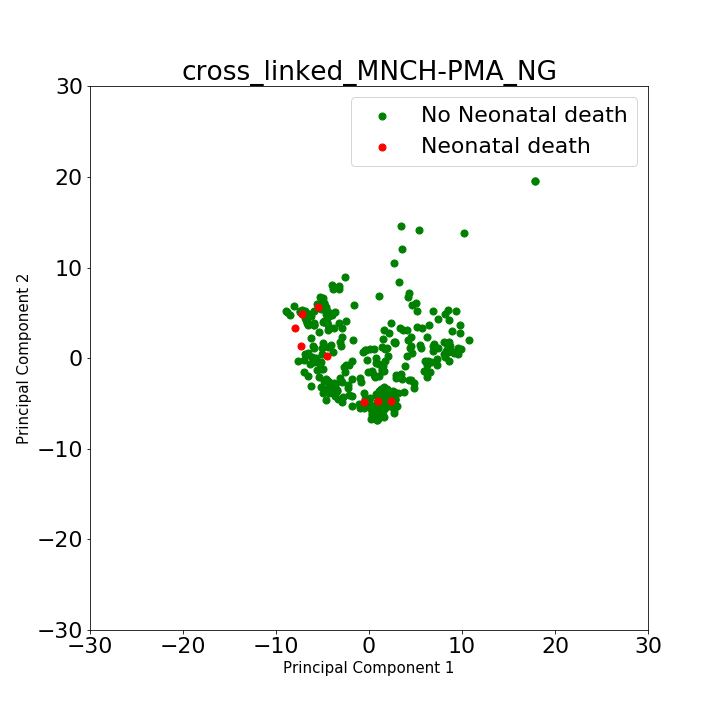}
    \caption{PCA projections of PMA data from Ethiopia (ET) before linkage is applied followed by the projections after PMA ET is linked with DHS of other african countries.}
    \label{fig:pcas}
\end{figure}

 %\vspace{-0.1cm}
% \paragraph{Datasets and performance metric:}
% \vspace{-0.1cm}
 We focus on the problem of neonatal death across Sub-Sahara African countries, and hence use datasets collected across two different efforts (DHS~\cite{dhs} and PMA~\cite{pma}) in the following countries: Burkina Faso (BF), Ethiopia (ET), Ghana (GH), Kenya (KE) and Nigeria (NG). 
%  The validation follows the linkage of PMA and DHS datasets from different countries as proposed, and prediction performance is compared to domain-specific prediction (i.e. datasets without linkage).
Due to the high imbalance between neonatal death and not (often ~2-4\%), we employed area under receiver operating characteristics (AUROC) as our performance metric. 
Results shown in Table~\ref{tab:my_label} demonstrate that the linkage mechanism helps to alleviate the prediction performance of neonatal death on PMA data of Ethiopia, when linked with DHS data of different countries, which were collected at different times and locations compared to the PMA. The PMA ET only achieves an AUROC of $45.9\%$ expectedly due to its small size and high degree of imbalance. However, this performance is improved to $90.0\%$ using the DHS ET, and improved further more to $94.1\%$ by using the DHS data of Ghana. The 2-D PCA projections in Fig.~\ref{fig:pcas} show the proposed linkage make the original data sparse  and hence eases prediction. 
% This validates the benefit of utilising cross-country datasets, even if they don't have overlapping data points between them.
\vspace{-0.1cm}
\begin{table}[]
    \centering
    \resizebox{0.8\linewidth}{!}{
    \begin{tabular}{l|ccccc}
    \hline 
    & \multicolumn{5}{c}{ Linked with DHS of:} \\
Linking method	&ET&	BF&	GH&	KE	& NG \\ \hline \hline
Random &	60.6 &	63.3 &	64.5 &	49.4 &	54.9 \\ \hline
Feature importance &	66.6 &	59.9 &	69.3 &	57.3 &	66.6 \\
Principal component analysis &	64.3 &	64.8 &	64.7 &	56.3 &	66.7 \\
Autoencoder &	90.0 &	90.0 &	\textbf{94.1} &	84.6 &	86.1 \\ \hline
    \end{tabular}
    }
    \caption{Increase in the neonatal death prediction on PMA data of Ethiopia when linked with DHS data of other African countries. Random linkage of samples is used as a baseline.}
    \label{tab:my_label}
    
\end{table}
%
%
% \section*{Discussion}
% \vspace{-0.4cm}
It has been shown that exploiting existing but disjoint datasets that were collected in different locations and times would help improve prediction performances in solving critical global healthcare challenges, where large and unified data collection is often difficult and expensive. The proposed data-level linkage of these datasets helps to achieve that, and importantly, can be extended to other sectors beyond healthcare. Example includes understanding and monitoring of multiple sustainable development goals by exploiting data collected across different streams.
% \section*{Acknowledgements}

% This work is funded by Bill \& Melinda Gates Foundation.

\nocite{langley00}
\newpage
\bibliography{ICML_ML4H_cross-domain}
\bibliographystyle{icml2020}

% \appendix
% \section{Do \emph{not} have an appendix here}

% \textbf{\emph{Do not put content after the references.}}
% %
% Put anything that you might normally include after the references in a separate
% supplementary file.

% We recommend that you build supplementary material in a separate document.
% If you must create one PDF and cut it up, please be careful to use a tool that
% doesn't alter the margins, and that doesn't aggressively rewrite the PDF file.
% pdftk usually works fine. 

% \textbf{Please do not use Apple's preview to cut off supplementary material.} In
% previous years it has altered margins, and created headaches at the camera-ready
% stage. 
%%%%%%%%%%%%%%%%%%%%%%%%%%%%%%%%%%%%%%%%%%%%%%%%%%%%%%%%%%%%%%%%%%%%%%%%%%%%%%%
%%%%%%%%%%%%%%%%%%%%%%%%%%%%%%%%%%%%%%%%%%%%%%%%%%%%%%%%%%%%%%%%%%%%%%%%%%%%%%%

\end{document}